\documentclass[graybox]{svmult}
\usepackage{graphicx}
\usepackage{caption}
\usepackage[section]{placeins}
\usepackage{url}
\usepackage[square,sort,comma,numbers]{natbib}
\usepackage{amsmath}
\makeatletter
\AtBeginDocument{%
  \def\doi#1{\url{#1}}}
\makeatother
\begin{document}

\title*{Reachability Analysis for Lexicase Selection via Community Assembly Graphs}
\titlerunning{Community Assembly Graphs}
\author{Emily Dolson\inst{1} \and Alexander Lalejini\inst{2}}

\institute{ Emily Dolson \at Michigan State University \email{dolsonem@msu.edu}
\and 
Alexander Lalejini \at Grand Valley State University, 
\email{lalejina@gvsu.edu}
}

\titlerunning{Community Assembly Graphs}
\authorrunning{Dolson and Lalejini}

% \keywords{lexicase selection, genetic programming, eco-evolutionary theory, multi-objective optimization}

\maketitle

\abstract*{Fitness landscapes have historically been a powerful tool for analyzing the search space explored by evolutionary algorithms. In particular, they facilitate understanding how easily reachable an optimal solution is from a given starting point. However, simple fitness landscapes are inappropriate for analyzing the search space seen by selection schemes like lexicase selection in which the outcome of selection depends heavily on the current contents of the population (i.e. selection schemes with complex ecological dynamics). Here, we propose borrowing a tool from ecology to solve this problem: community assembly graphs. We demonstrate a simple proof-of-concept for this approach on an NK Landscape where we have perfect information. We then demonstrate that this approach can be successfully applied to a complex genetic programming problem. While further research is necessary to understand how to best use this tool, we believe it will be a valuable addition to our toolkit and facilitate analyses that were previously impossible.}

\abstract{Fitness landscapes have historically been a powerful tool for analyzing the search space explored by evolutionary algorithms. In particular, they facilitate understanding how easily reachable an optimal solution is from a given starting point. However, simple fitness landscapes are inappropriate for analyzing the search space seen by selection schemes like lexicase selection in which the outcome of selection depends heavily on the current contents of the population (i.e. selection schemes with complex ecological dynamics). Here, we propose borrowing a tool from ecology to solve this problem: community assembly graphs. We demonstrate a simple proof-of-concept for this approach on an NK Landscape where we have perfect information. We then demonstrate that this approach can be successfully applied to a complex genetic programming problem. While further research is necessary to understand how to best use this tool, we believe it will be a valuable addition to our toolkit and facilitate analyses that were previously impossible.}

\section{Introduction}

Lexicase selection is a state-of-the art parent-selection algorithm for genetic programming \citep{spectorAssessmentProblemModality2012}. It has proven highly effective across a wide variety of problems~\citep{moore_lexicase_2017,pappa_faster_2023,la_cava_epsilon-lexicase_2016,ding_optimizing_2022,lalejini_artificial_2022,metevier_lexicase_2019}, and has spawned many variants \citep{lacavaEpsilonLexicaseSelectionRegression2016, spectorRelaxationsLexicaseParent2018, hernandezRandomSubsamplingImproves2019, boldiInformedDownSampledLexicase2023}. One challenge of working with lexicase selection, however, is that most fitness-landscape-based analytical techniques do not directly apply to it. Fitness landscapes represent the mapping of genotypes to fitness and the adjacency of genotypes to each other, providing intuition for which genotypes are (easily) reachable from which other genotypes via an evolutionary process. Because lexicase selection is designed for scenarios where multiple factors (e.g. different test cases that evolved code is run on) determine solution quality, there is no single fitness landscape for a given problem in the context of lexicase selection (for further details on lexicase selection works, see Section \ref{sec:lex-background}).

Moreover, a candidate solution's probability of being selected by lexicase selection depends entirely on the composition of the population of other solutions that it is competing with \citep{lacavaProbabilisticMultiObjectiveAnalysis2018}. In other words, ecological dynamics play a large role in lexicase selection \citep{dolson_ecological_2018, dolsonApplyingEcologicalPrinciples2018}; the fitness landscape for any individual criterion is constantly shifting due to endogenous change. Thus, even if we could calculate individual criterion fitness landscapes, there would not be a meaningful way to combine them into a single static model predicting the population's change over time. 

In cases where evolutionary algorithms fail to solve problems, fitness-landscape-based analyses are useful for understanding why \citep{ostmanPredictingEvolutionVisualizing2014}. Although alternative analytical techniques can fill some of this gap \citep{hernandez_suite_2022}, it would be useful to have a technique for conducting reachability analysis, i.e. identifying whether lexicase selection is capable of reaching a solution to a given problem under a given configuration. In particular, such an analysis would help distinguish instances where lexicase selection simply needs more time or a larger population size from instances where it is incapable of or unlikely to solve a problem.

A technique capable of supporting reachability analysis for lexicase selection would need to consider 1) what populations of solutions can exist, and 2) what additional solutions are reachable from a given population. If we have (1), (2) is relatively easy to calculate, as it simply requires identifying solutions mutationally adjacent to those in the population. Identifying possible populations, however, may at first seem intractable. 

Fortunately, this problem has been solved by ecologists seeking to predict the way that an ecological community will change over time. Here, we propose to use a tool from ecology -- the community assembly graph -- to improve our understanding of the adaptive ``landscape'' experienced by lexicase selection. This approach can likely be generalized to other evolutionary algorithms with strong ecological interactions between members of the population.

%When a given algorithm fails to find a solution to a problem, fitness landscapes can provide insight into why. 

%It is a state-of-the-art approach to genetic programming, and has also demonstrated success in other areas of evolutionary computation \citep{aenuguLexicaseSelectionLearning2019, ding2022optimizing}.  

%Consequently, traditional fitness-landscape based analysis is challenging. 

%For each selection event, all $N$ of these factors are placed in a random order. The algorithm then iterates through these factors, eliminating all but the best-performing solutions on each until only one solution remains. The presence of multiple fitness criteria poses a an obvious problem to fitness landscape analysis: instead of optimization occurring on a single landscape, it occurs on some amalgamation of $N$ landscapes.

%Despite its power, theoretical work done on lexicase selection has been limited \citep{lacavaProbabilisticMultiObjectiveAnalysis2018, helmuthPopulationDiversityLeads2022b, dolsonCalculatingLexicaseSelection2023}. This lack of research may be due, in part, to the fact that lexicase selection defies many standard fitness landscape analysis techniques. 

\section{Approach}

\subsection{Community Assembly Graphs}

\begin{figure}
    \centering
    \includegraphics[width=0.6\linewidth]{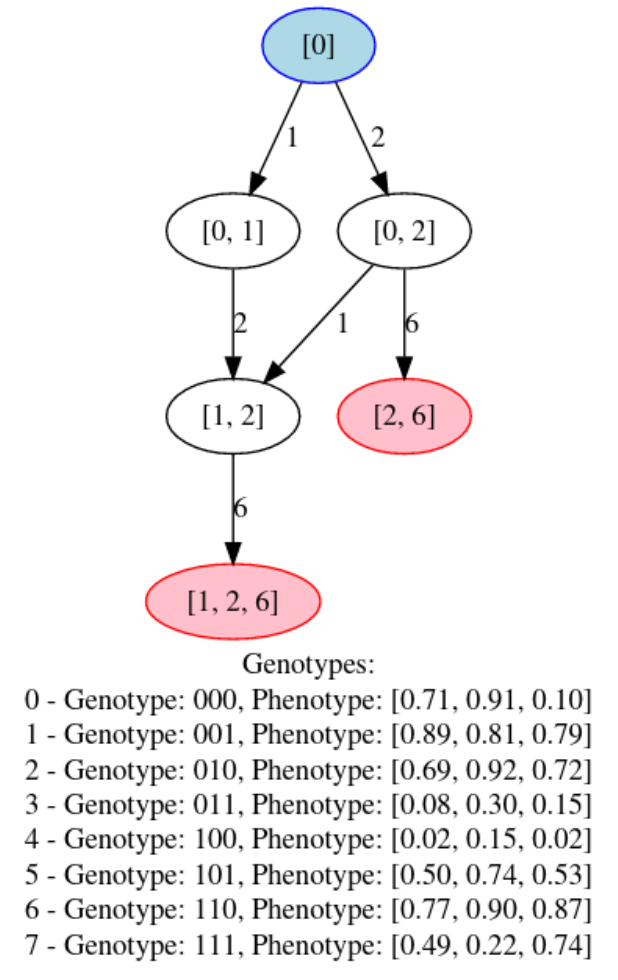}
    \caption{\textbf{A simple community assembly graph}. This graph represents lexicase selection on a representative NK fitness landscape with N=3, K=2. The fitness contributions of each of the three genes function as the three fitness criteria for lexicase selection. Node labels indicate the ids of genotypes that are present in the community represented by each node. The genotype and phenotype corresponding to each id are shown below the graph. Edge labels indicate which genotype was added to the community to cause a transition from the edge's source node to its destination node. Here, for simplicity, evolution is arbitrarily assumed to have started from a population containing only genotype 0. The two mutationally adjacent genotypes to 0 are 1 and 2, so we consider the effect of adding either of them to the starting community. Both genotype 1 and genotype 2 can stably coexist with genotype 0 (they each have a fitness criterion on which they outperform genotype 0 and a fitness criterion on which they do not), so adding either one results in a two-genotype community. Ultimately, there are two different sink nodes in this graph; evolution will likely stagnate when it reaches either of them. Because genotype 1 only appears in one of the sink nodes, we can conclude that it will probably not always be found, despite having the highest score on objective 1.}
    \label{fig:example_cag}
\end{figure}

In ecology, the process of building up a collection of ``species'' that can coexist with each other is called community assembly. There is a vast literature of ecological theory on this process that may be relevant to lexicase selection. For now, we will borrow a single concept from this literature: community assembly graphs \citep{servan2021tractable, hang1993assembly} (see Figure \ref{fig:example_cag}). In these graphs, each node represents a possible community composition (i.e. set of species that could stably coexist with each other in the same space). Community compositions can be represented as bit-strings where each position corresponds to a species that could potentially be present. A one indicates that that species is present, while a zero indicates that it is absent. While this technique may potentially make for a very large graph ($2^n$ nodes, where $n$ is the number of possible species), note that the graph only includes stable communities. Many collections of species cannot all coexist with each other for more than a few generations and thus are not represented as nodes in the graph. Edges between nodes are directed and indicate transitions that can occur via the addition of a new species to the community at the source node.

Note that in the previous paragraph we referred to ``species'', as these are the units that can belong to a community in ecology. In evolutionary computation, however, we do not usually attempt to define species. Instead, we can choose an alternative unit of organization to compose our communities out of. One obvious option would be genotypes (i.e. representations of the entire genome). However, we recommend using phenotypes (i.e. representations of a solution's selection-relevant traits) instead, as all genotypes that share the same phenotype will also be ecologically equivalent. In the context of lexicase selection, we define ``phenotypes'' to be the vector of scores on all fitness criteria/test cases (sometimes referred to as ``error vectors'' in the literature).

Another important difference between community assembly graphs in ecology and the approach we propose here concerns which phenotypes may be added to a community. In traditional ecological scenarios, it is assumed that new species are being introduced to a local community from some nearby location, and thus any species could be introduced to any community. In evolutionary computation, however, new phenotypes come instead from the process of evolution. Consequently, for the purposes of making community assembly graphs for lexicase selection, we only consider the introduction of phenotypes that are mutationally-adjacent to a phenotype that is already present in a community. For a very simple example community assembly graph for lexicase selection, see Figure \ref{fig:example_cag}. 

A run of lexicase selection can be approximately modeled as a random walk on this community assembly graph, although in practice some paths will have higher probabilities of being taken than others. Sink nodes in this graph indicate possible end states for the process of evolution by lexicase selection on the problem being solved. By definition, a sink node is a community that has no mutationally-adjacent solutions capable of ``invading'' (i.e. surviving in) it. Thus, it is impossible to escape (setting aside, for now, the possibility of multiple simultaneous mutations). Any evolutionary scenario that produces multiple sink nodes has multiple possible outcomes. A corollary of this observation is that any solution not appearing in all sink nodes is not guaranteed to be found, and that running evolution for longer will not change that fact.

\subsection{Calculating stability}

As the nodes in the community assembly graph correspond to stable communities, identifying which communities are stable is a critical step in calculating a community assembly graph. We define stable communities as follows:

\begin{definition}
A \textit{stable community} is a population of phenotypes such that, if no mutations were allowed, the set of phenotypes present would have a high probability of remaining the same through $G$ generations worth of selection events.
\end{definition}

In other words, a stable community is a set of phenotypes expected to stably coexist for $G$ generations. Note that this definition means stable communities can only be defined with respect to fixed values of $G$, population size, and some threshold for what counts as a ``high'' probability of survival. 

What is the purpose of allowing generations to vary by including the $G$ term? In genetic programming, each phenotype can be represented as a neutral network of equivalent genotypes that are mutationally-adjacent to each other \citep{banzhaf_2023_neutrality}. While some phenotype $A$ may be adjacent to some other phenotype $B$, it is unlikely that all genotypes within $A$ are adjacent to all genotypes in $B$. Instead, to discover phenotype $B$ from phenotype $A$, evolution likely needs to traverse the neutral network of phenotype $A$. Consequently, it would be incorrect to assume that $B$ can be discovered as soon as $A$ is in the population. Instead, $A$ must first be able to survive for some amount of time related to the topology of its neutral network. If there is no neutrality in a landscape, $G$ can simply be set to 1.

% It allows us greater flexibility in our representation of solutions. If solutions are all represented as genotypes, we can simply set $G=1$, as surviving for a single generation is sufficient to provide a given genotype with the opportunity to have offspring with mutations that create new genotypes. However, the space of possible genotypes for most genetic programming problems is intractably large. Moreover, in lexicase selection only the phenotype (performance on test cases) matters for the purposes of community stability. Consequently, we can dramatically simplify our analysis by representing all solutions as phenotypes. However, if solutions represent phenotypes, the assumption that a solution only needs to survive for a single generation in order to produce offspring with a new phenotype becomes inaccurate. Fitness landscapes often have substantial neutral spaces corresponding to a single phenotype that must be traversed before a new phenotype can be discovered. Thus, when analyzing evolution on this kind of fitness landscape, we can use a higher value of $G$ to ensure that we are only considering communities that are stable long enough for new phenotypes to evolve.

We can identify stable communities in lexicase selection using a sequence of two equations. First, we identify $P_{lex}$, the probability of each member of the population being selected in a given selection event. $P_{lex}$ can be calculated with the following equation, derived in \citep{lacavaProbabilisticMultiObjectiveAnalysis2018}:

\begin{align}
    P_{lex}(i | Z, N) =  \begin{cases}
    1 & if | Z | = 1 \\
    1/|Z| & if |N|=0 \\
    \frac{\sum\limits_{j = 0}^{|N|} P_{lex}(i, \{z \in Z | z \textit{ elite on } N_j \},  \{n_i \in N | i != j \})}{|N|} & else
    \end{cases}
\label{eq:plex}
\end{align}

Note that the run-time of this function is exponential, as the problem of calculating $P_{lex}$ is $NP$-Hard \citep{dolsonCalculatingLexicaseSelection2023}. However, in practice $P_{lex}$ can be calculated fairly efficiently using a variety of optimizations \citep{dolsonCalculatingLexicaseSelection2023}.

Once we have calculated $P_{lex}$, we can calculate the probability of each phenotype surviving for $G$ generations, $P_{survival}$, based on the following equation (adapted from \citep{dolson_ecological_2018}):

\begin{equation}
    P_{survival}(i, S, G, pop) = (1 - (1 - P_{lex}(i, pop))^S)^G
\label{eq:p_survival}
\end{equation}

Where $i$ is the individual $P_{survival}$ is being calculated for, $S$ is the population size, $G$ is the number of generations, and $pop$ is the current population. As $S$ and $G$ get large, this function begins to approximate a step function \citep{dolson_ecological_2018}. Consequently, we can often safely make the simplifying assumption that the survival of each member of the population is either guaranteed or impossible.

\subsection{Assumptions}

Three important simplifying assumptions are made in the construction of community assembly graphs in this paper. The first is that it is only possible to generate single-step mutants during a reproduction event. Obviously, many mutation schemes would violate this assumption. If necessary, it would be easy to amend our graph construction process to consider larger numbers of mutations. We refrain from doing so here in the interest of improved tractability. Whether this assumption reduces the utility of community assembly graph analysis is an important open question.

The second simplifying assumption is that no more than one new phenotype is added to the population per generation. In some cases, the result of two phenotypes being added simultaneously may be different than the results of those phenotypes being added in sequence. In such cases, the community assembly graph model may incorrectly predict the behavior of lexicase selection. In theory, if the landscape has sufficient neutrality (and thus $G$ should be large), it should be relatively uncommon for these cases to occur. Whether these cases significantly impact the behavior of lexicase selection is another open question. Moreover, even if these events are important, community assembly graphs give us a framework for quantifying that importance.

The third simplifying assumption is that the population size of each phenotype is irrelevant and we can conduct analyses purely on the basis of presence vs. absence of each phenotype. This assumption is likely justified, as population size will only impact $P_{lex}$ in scenarios where multiple solutions tie. Again, empirical analysis is warranted to determine whether these scenarios play a significant role in lexicase selection's behavior.

\subsection{Reachability analysis}

Analyzing the topological properties of the community assembly graph can provide a variety of insights into the landscape being explored by lexicase selection. For now, we will focus on a single one: reachability. Reachability analysis asks whether a path exists from a given starting point, $A$, to a given ending point, $Z$. Thus, it allows us to determine, for an initial population ($A$), whether lexicase selection is capable of finding an optimal solution ($Z$). Using a community assembly graph, we can conduct this analysis via a simple graph traversal. We can also ask a stronger question about reachability: are there paths from $A$ that do not ultimately lead to $Z$? Community assembly graphs also make this form of reachability easy to assess; we need only determine whether there are any sink nodes (i.e. nodes with no outgoing edges) other than $Z$ that are accessible from $A$. The answer to this question tells us approximately how likely lexicase selection is to find an optimal solution to a given problem from a given initial population. 

\subsubsection{Tractability}

Why do we start from a predefined initial population, $A$? This restriction keeps the problem substantially more tractable. The full community assembly graph for a given configuration of lexicase selection on a given problem would have as many as $2^n$ nodes, where $n$ is the number of unique phenotypes that could possibly exist. While many of these communities can be excluded from the graph due to being unstable, the set of unique performance vectors is already potentially uncountable. By starting from a defined point, we can avoid attempting to enumerate all possible vectors. Fortunately, in the context of genetic programming, it is often reasonable to assume that we start from a population containing a single error vector: the one representing minimal scores on all fitness criteria.

However, in many practical cases, this restriction is still insufficient to make reachability analysis tractable. In these cases, we can limit our analysis to an even smaller sub-graph ($G$) containing only the nodes that we are most likely to discover. We can find $G$ using a modified version of Dijkstra's algorithm. Like Dijkstra's algorithm, our algorithm is a graph traversal in which all newly discovered nodes are placed in a priority queue. On every iteration, we remove the top node from the priority queue and ``explore'' it, \textit{i.e.} we place all undiscovered nodes that it has an edge to in the priority queue. However, whereas Dijkstra's algorithm assigns priorities based on minimal summed path length, our algorithm will assign priorities based on the hitting probability of each node given the portion of $G$ that we have explored so far. Hitting probability is the probability that a node will be visited in the course of a random walk along the graph. Note that it differs from the more commonly used properties of hitting time (the average time it will take a random walk to reach a node) and PageRank (the probability of ending a random walk on a node). We will denote the hitting probability for node $i$ when starting from node $A$ as $H_{iA}$.

Using this traversal algorithm, we can explore the top $N$ communities that are most likely to occur at some point during an evolutionary process started from the community represented by node $A$. To maintain tractability, we stop our traversal after we have fully explored N nodes. As is standard for graph traversal algorithms, we consider a node fully explored once all nodes that it has a directed edge to have either been explored or are in the priority queue of nodes to explore in the future. An interesting side effect of this algorithm is that at the end, we are left with nodes with three different states: 1) fully explored, 2) discovered but unexplored, and 3) undiscovered. Only fully explored nodes will be included in $G$ for follow-up analysis. However, we can use our knowledge of the nodes in state 2 to determine which nodes in $G$ truly have no outgoing edges and thus should be considered sink nodes for the purposes of reachability analysis (as opposed to nodes in $G$ that have outgoing edges to nodes with too low a hitting probability to be included in $G$).

\subsubsection{Calculating hitting probability}

Note that calculating hitting probability is non-trivial, particularly when we are calculating it as the graph is being traversed. Whenever we discover a new path to a node, its hitting probability (and thus its priority in the priority queue) must be updated\footnote{While standard heap-based priority queues do not support updating priorities, a Fibonacci heap can be used to enable this operation in constant time.}. In the case of simple paths, these updates can be handled by summing the probabilities of each path. However, when a cycle is discovered, things become more challenging, as we have effectively just discovered an infinite number of new paths to all nodes that are successors (direct or indirect) of any node in the cycle.

As a simplifying assumption for the purposes of this proof-of-concept paper, we ignore the effect of cycles on our priorities. Consequently, we may underestimate the hitting probabilities of some nodes. As we expect cycles to be fairly rare in the context of lexicase selection (and, when they do occur, they are likely evenly distributed across the graph), we do not anticipate this bias has a dramatic impact on our analyses here.

Development of an algorithm to appropriately handle cycles is underway as part of follow-up research. One approach to solving this problem is to employ a combination of infinite series and combinatorics to calculate the probabilities of taking each possible path through the graph. Unfortunately, this approach is challenging to implement, due to the possibility of many overlapping cycles. A more straightforward approach is to reduce the problem of calculating hitting probability to the much easier problem of calculating PageRank. This reduction can be achieved by constructing a modified version of $G$, which we will term $G'_i$, for each node $i$ that we want to calculate the hitting probability of. $G'_i$ is identical to $G$ except that all outgoing edges from node $i$ are removed. Consequently, any path in $G$ that would include node $i$ corresponds to a path in $G'_i$ that ends at node $i$.

One other aspect of our hitting probability calculations that it is important to be aware of is that all probabilities we calculate are conditional on the part of the graph already traversed at any point in time. It is possible that at the time we reach our stopping condition of having explored $N$ nodes, there will be nodes in the priority queue that have a higher hitting probability than nodes that we have already explored. Similarly, our analysis makes no prediction about what will happen in the unlikely event that evolution reaches a node not contained in $G$. It would be worthwhile in future work to explore the impact of these shortcomings and whether they can be overcome. 

\section{Background}

For readers interested in slightly more context on the inner workings of lexicase selection and the history of community assembly graphs, we offer some additional context.

\subsection{Lexicase Selection}
\label{sec:lex-background}

Lexicase selection is a parent selection algorithm designed to operate in contexts where multiple criteria affect fitness \citep{spectorAssessmentProblemModality2012}. In genetic programming, these criteria are generally the candidate solution's performance on various test cases. To select an individual from the population, the test cases are placed in a random order and the entire population is placed into a pool of solutions eligible for selection. The algorithm then iterates through the test cases in their randomized sequence. For each test case, the best performing solutions in the pool are identified. These solutions are allowed to remain in consideration for selection, while all others are removed from the pool. The last remaining solution is selected. If there is a tie, a random solution is selected.

\subsection{Community Assembly Graphs}

A central set of questions in community ecology concern the extent to which ecological communities assemble in predictable ways. As an obvious way to systematically address this class of question, community assembly graphs have a long history in ecological theory \citep{hang-kwangAssemblyEcologicalCommunities1993, schreiberSimpleRulesCycling2004, capitanStatisticalMechanicsEcosystem2009, servan2021tractable}. However, in the context of ecology there are a number of factors that interfere with using community graphs to accurately model community dynamics. Servan and Allesina identify three specific complicating factors: rate of invasion, size of invasion, and timing of invasion \citep{servan2021tractable}. Uncertainty in these factors makes it hard to predict the outcome of invasions \textit{a priori}. 

In evolutionary computation, rate of invasion translates to the rate of discovery of new phenotypes. As previously discussed, due to the neutrality inherent in most genetic programming problems, we can assume this is relatively slow. Moreover, we can tune the rate at which we expect to discover new phenotypes using the $G$ parameter. For this same reason, we can be relatively confident that the size of the invasion is always a single individual with the new phenotype. Interestingly, these are the same simplifying assumptions Servan and Allesina end up making \citep{servan2021tractable}, suggesting that evolutionary computation is well-suited to this framework.

Timing of invasion is the most challenging of these factors to account for in the context of evolutionary computation. Specifically, the major variable is how the invasion is timed relative to fluctuations in other phenotype's population sizes. Fortunately for our immediate purposes, it is unlikely to make a large difference in the context of lexicase selection, as the ecology of lexicase selection is not heavily influenced by population size \citep{dolson_ecological_2018}. However, this challenge will be important to consider when using community assembly graphs to analyze other selection schemes. Servan and Allesina handle this problem by assuming populations are at equilibrium, which may be a viable approach in evolutionary computation as well.

\section{Proof of Concept in NK Landscapes}

To confirm that community assembly graphs accurately predict the dynamics of lexicase selection, we initially test them out in the context of a simple and well-understood problem: NK Landscapes \citep{kauffmanGeneralTheoryAdaptive1987}. 

\subsection{Methods}

NK landscapes are a popular and well-studied class of fitness landscapes with a tunable level of epistasis. They are governed by two parameters: N and K. Genomes that evolve on NK Landscapes are bit-strings of length N. Each of the N sites contributes some amount to the overall fitness of the bitstring, depending on its value and the value of the K adjacent sites. Fitness contributions are calculated based on a randomly generated lookup table for each site, which contains a different fitness contribution for every possible value that the genome could have at that site.

For example, if K=0, each site would have two possible fitness contributions: one which would be contributed if the bit at that position were a 1 and one which would be contributed if it were a 0. If K=1, there would be 4 possible values (for 00, 01, 10, or 11), and so on. For the purpose of lexicase selection, the fitness contribution at each site is treated as a separate fitness criterion.

\begin{figure}
    \centering
    \includegraphics[width=\linewidth]{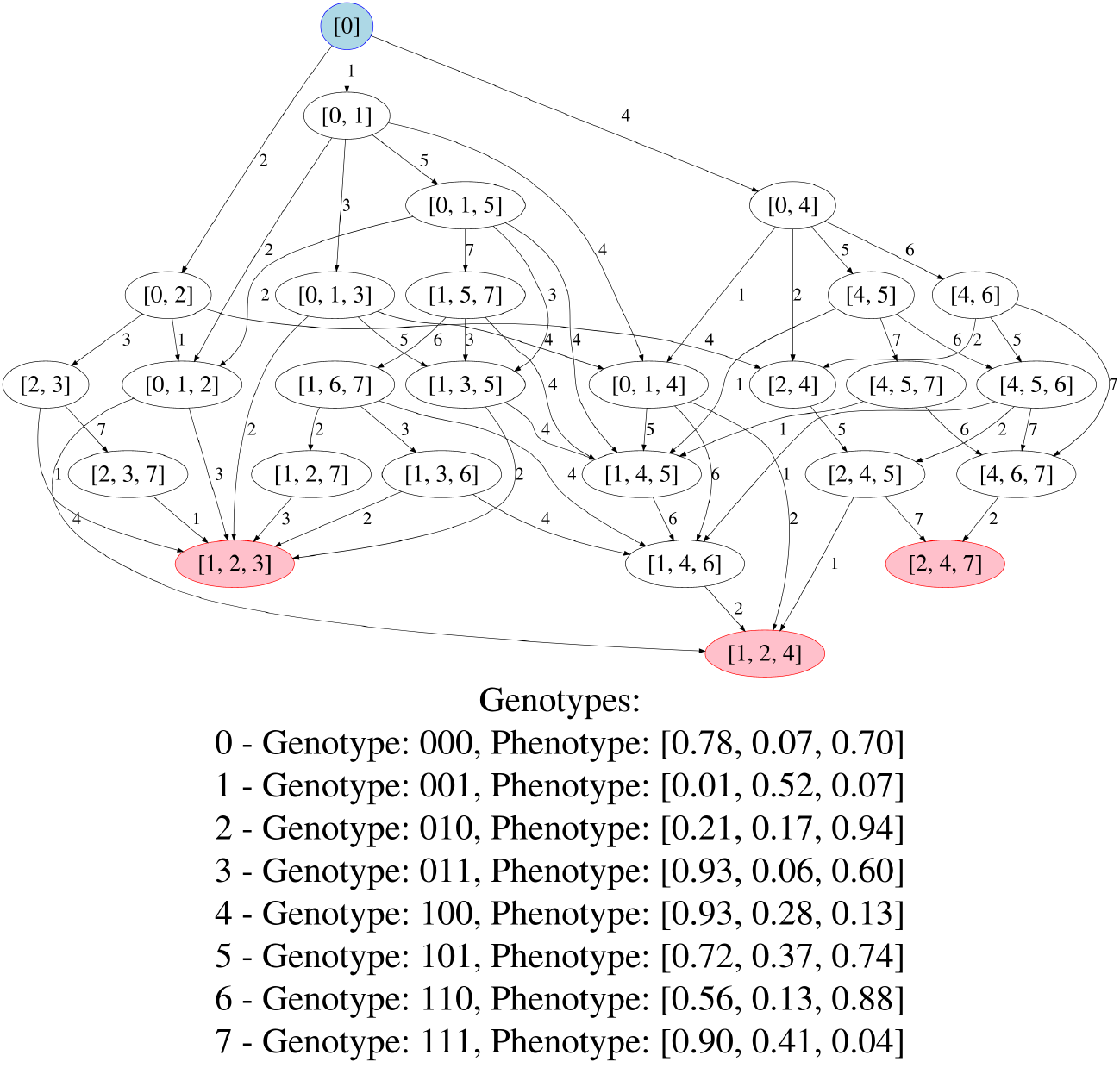}
    \caption{Community assembly graph for the NK fitness landscape used in these experiments. N=3, K=2. The starting population is assumed to contain only the bitstring 000. Starting node is blue, sink nodes are red.}
    \label{fig:complex_cag}
\end{figure}

We generate a community assembly graph for an arbitrary NK fitness landscape (see Figure \ref{fig:complex_cag}). We chose a small value of N (3) for the purposes of being able to visualize the entire community assembly graph and a relatively high value of K (2) to increase the complexity of the search space. In calculating this community assembly graph, we assume $G=1$ and population size is 100.

To assess the accuracy of the community assembly graph's predictions, we then ran 30 replicate runs of evolution on this NK Landscape. These runs were allowed to proceed for 500 generations with a population size of 100. We tested two different mutation rates. The first was low (.001 per-site) to ensure that the assumptions of the community assembly graph were not violated. The second was more realistic (.1 per-site), to test the impact of violations to the assumptions.

\subsection{Results}

At the low mutation rate, the community assembly graph perfectly identified the set of final communities observed (see Table 1). At the higher mutation rates, the population consistently converged on a single community. This community was one of the possible end states predicted by the community assembly graph.

\begin{table}[]
\centering
\begin{tabular}{|c|c|c|c|c|}
\hline
\textbf{Mutation Rate} & \textbf{1,2,3} & \textbf{1,2,4} & \textbf{2,4,7} & \textbf{Other} \\ \hline
\textbf{.001}          & 11             & 13             & 6              & 0              
\\ \hline
\textbf{.01}            & 0              & 30             & 0              & 0      
\\ \hline
\textbf{.1}            & 0              & 30             & 0              & 0              \\ \hline
\end{tabular}
\vspace{1em}
\caption{Final communities reached by lexicase selection on the example NK Landscape.}
\end{table}

These preliminary results suggest that violations to the assumptions of the community assembly graph may alter the probabilities of observing the different final communities. Further experiments to determine whether violations to the assumptions can ever lead to convergence on non-sink nodes are underway (although such an occurrence should theoretically be impossible).

Encouragingly, the community that was consistently reached under the high mutation rates is the ``better'' community to find, in that it includes the best-performing solutions on each fitness criterion. We hypothesize that this trend may generalize across evolutionary scenarios, and thus that predictions based on community assembly graphs may represent a near worst-case scenario. While more work is needed to test this hypothesis, the reasoning behind it is as follows: the primary assumption that is violated by having a high mutation rate is the assumption that only one new ``species'' is introduced at once. Violating this assumption should tend to favor the best-performing solutions, as they will tend to outcompete other solutions that are introduced simultaneously.

\section{Proof of Concept in Genetic Programming}

While the results on NK Landscape are promising, they are dramatically simpler than realistic genetic programming problems. To understand whether community assembly graphs are a practical tool for genetic programming, we next test them on a selection of problems from the first and second program synthesis benchmark suites~\citep{helmuth_general_2015,helmuth_psb2_2021}.

\subsection{Methods}

Any community assembly graph of an evolutionary algorithm is inherently specific to a specific problem, genetic representation, mutation scheme, and selection scheme. Here, we use community assembly graphs to conduct a supplemental analysis of the experiments in~\citep{lalejini2023phylogenyinformed}.

These experiments used SignalGP as a genetic representation~\citep{lalejini_evolving_2018}. SignalGP is a tag-based linear genetic programming system in which programs evolve sequences of instructions to execute in response to receiving external signals. Instructions can take arguments which indicate which data they operate on.  Mutations can occur in four forms: 1) one instruction is replaced with a different instructions, 2) an instruction is inserted into the genome, 3) an instruction is deleted from the genome, or 4) the arguments to an instruction are changed. For more details, see~\citep{lalejini_evolving_2018}. Although Lalejini et. al \citep{lalejini2023phylogenyinformed} explored multiple variations on lexicase selection, here we will focus only on standard lexicase selection. To carry these analyses out for other selection schemes in the future, we would need to adjust Equation \ref{eq:plex} accordingly.

It is not tractable to map out the entire mutational landscape of SignalGP. Instead, we must sample a representative portion of the landscape. A representative sample should, theoretically, include the regions of genotype space where we expect evolution to end up. We propose that the best way to find these regions is to conduct multiple replicate runs of evolution in the scenario of interest. Here, we use 10 replicate runs per condition. From each of these runs, we extract the full genotype-level phylogeny of the final population i.e. the full ancestry tree of which genotypes descended from which other genotypes ~\citep{dolson_phylotrackpy_2023}. These phylogenies will indicate what parts of the genotype space evolution ultimately traversed.

However, that information alone is insufficient, as we must also know the ecological context that each genotype is likely to find itself in. To obtain this additional information, we conduct mutational landscaping analysis around each genotype in each phylogeny. For each genotype, we produce 10,000 random mutants and test their performance on all test cases. We record all instances where a mutant had a different phenotype (i.e. test case performance profile) than the genotype it was generated from. By aggregating this data, we produce a network showing the probability of mutating from each phenotype to each other phenotype.

Finally, we construct the community assembly graph. Phenotype, rather than genotype, determines the ecological impact of a given solution being present in the population. Thus, each node in the community assembly graph represents a combination of phenotypes. Each community has the potential to have a given phenotype introduced to it if and only if that phenotype is mutationally adjacent to one of its member-phenotypes. The probability of mutating from one phenotype to another determines the probability of that mutation occurring.

Even with this sampling approach, the full community assembly graph is intractably large. To avoid this problem, we use the probability-based priority queue approach discussed in section 2.4.1. For the purposes of comprehensible data visualization, we explore a relatively small (100) number of nodes per graph.

\subsection{Results}

\begin{figure}
    \centering
    \includegraphics[width=\linewidth]{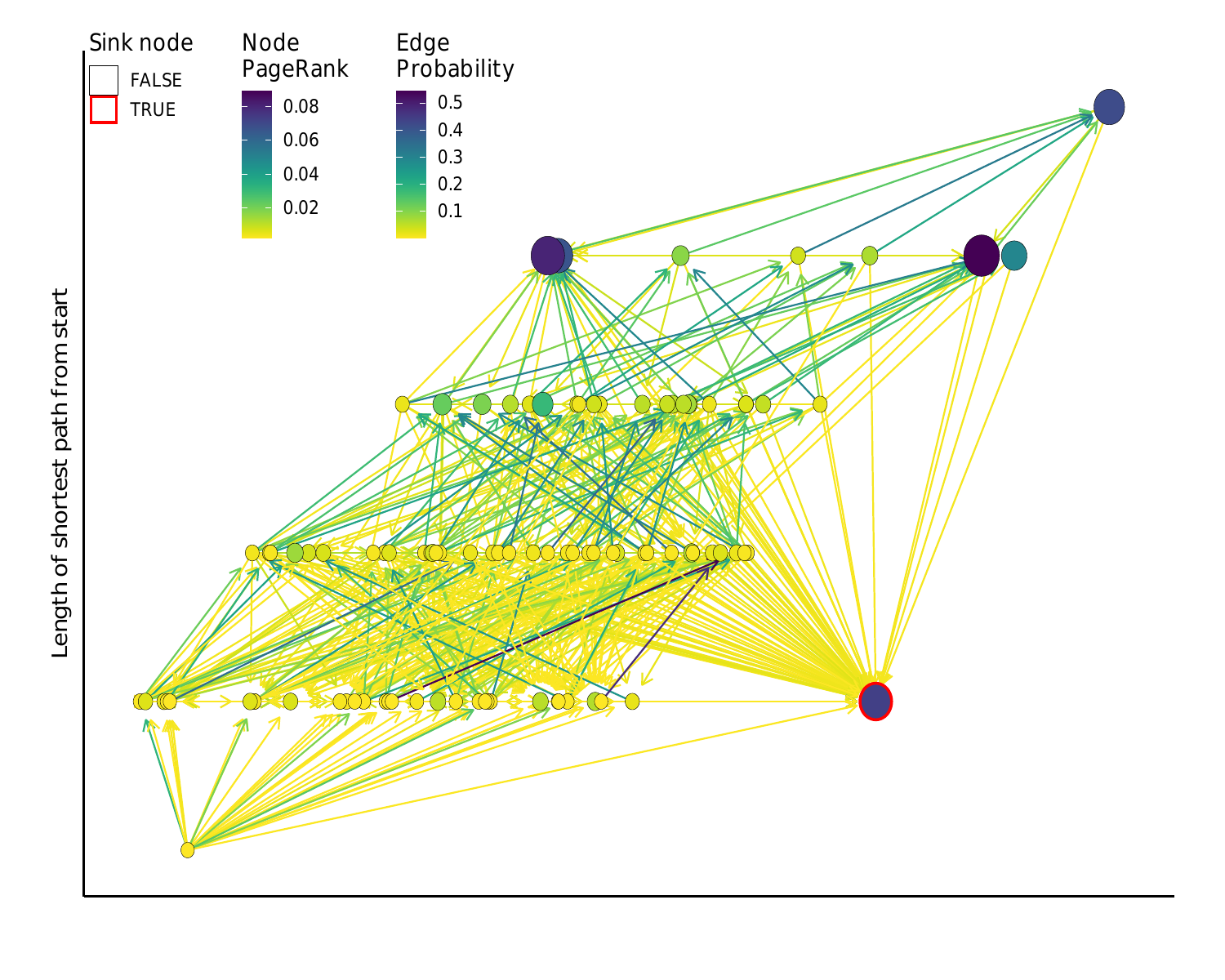}
    \caption{\textbf{Community assembly graph of the 100 most accessible communities for the grade problem.} Node colors and sizes indicate the PageRank of each node, which translates to the probability of a random walk ending on each node. Edge colors indicate the probability of choosing each edge. Nodes are arranged along the y axis according to how far away from the starting node they are (measured as shortest path). The starting node (representing a community containing only the worst-performing phenotype) is the lowest node on the y axis. Note that this portion of the graph contains only one true sink node (outlined in red). In this case, that node represents the optimal solution, indicating that the solution for this problem is indeed reachable.}
    \label{fig:grade}
\end{figure}

We first conduct a community assembly graph for the grade problem~\citep{helmuth_general_2015} (see Figure \ref{fig:grade}). Lalejini et. al found that this problem was solved fairly consistently~\citep{lalejini2023phylogenyinformed}, suggesting that it is relatively easy for SignalGP to solve. Indeed, the perfect solution appears in this graph and is the only sink node. Thus, it is reachable from our starting point (the community containing only the phenotype with scores of 0 for all test cases), and evolution is unlikely to get stuck anywhere else in between. The community corresponding to the perfect solution also has a relatively high PageRank, suggesting we should usually expect evolution to reach it.

Interestingly, there is an edge directly from the starting community to the optimal solution. However, the probability of taking this path is very low. Likely, the existence of this edge is due to mutational landscaping around the optimal genotype; it is often fairly easy to get a mutation that completely breaks a good solution. Since all mutations are bidirectional, the existence of such mutations means that it is technically also possible to mutate from the starting phenotype directly to the ending phenotype. Nevertheless, it is important to be aware that the probability of starting with a \textit{genotype} that makes such a mutation possible is vanishingly small.

\begin{figure}
    \centering
    \includegraphics[width=\linewidth]{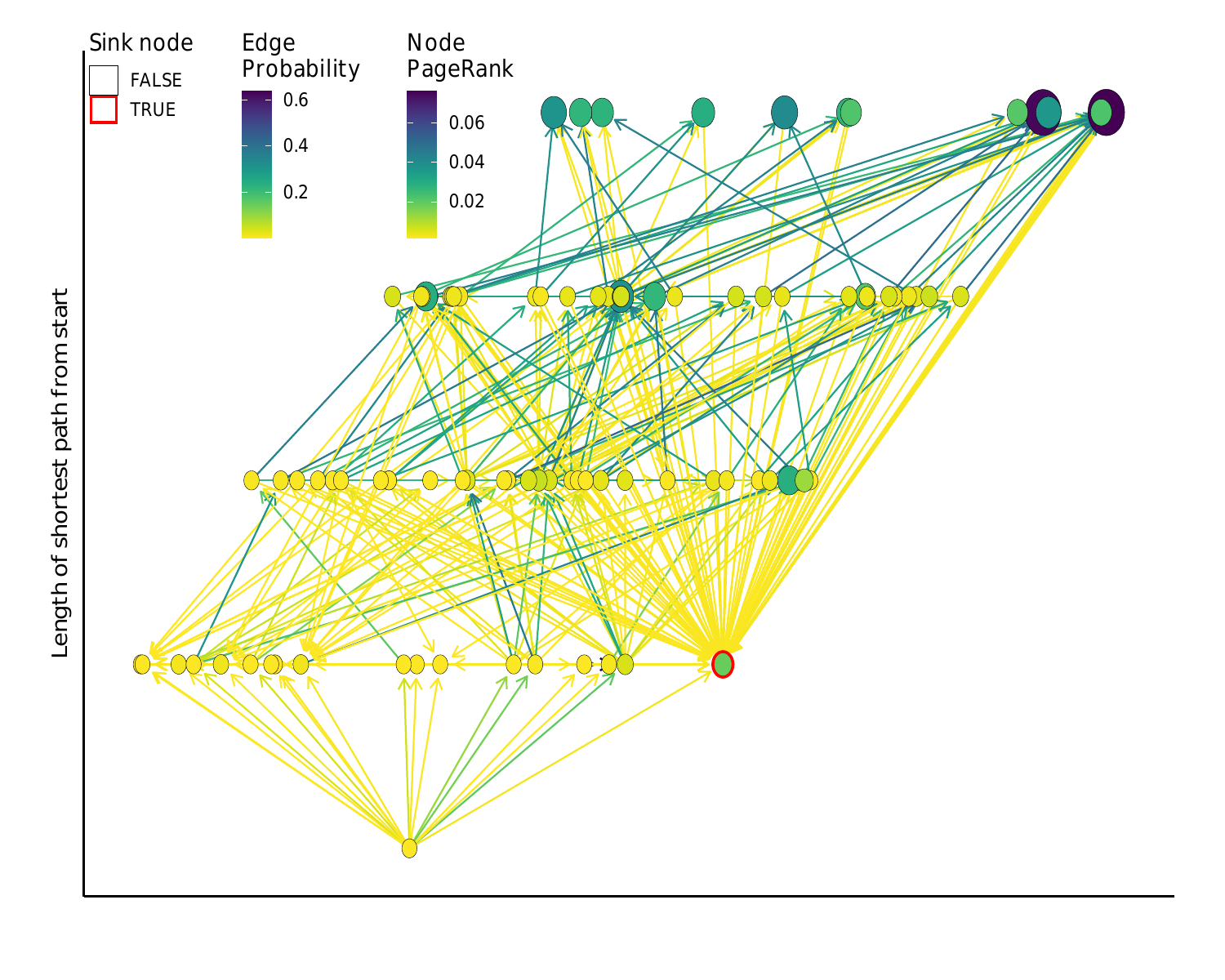}
    \caption{\textbf{Community assembly graph of the 100 most accessible communities for the median problem.} Node colors and sizes indicate the PageRank of each node, which translates to the probability of a random walk ending on each node. Edge colors indicate the probability of choosing each edge. Nodes are arranged along the y axis according to how far away from the starting node they are (measured as shortest path). The starting node (representing a community containing only the worst-performing phenotype) is the lowest node on the y axis. This portion of the graph contains only one true sink node (outlined in red), which represents the optimal solution.}
    \label{fig:median}
\end{figure}

The community assembly graph for the median problem~\citep{helmuth_general_2015} is qualitatively similar (albeit with the optimal solution having a somewhat lower PageRank) (see Figure\ref{fig:median}). Since Lalejini et. al found similar performance between the grade and median problems~\citep{lalejini2023phylogenyinformed}, the similarity of these graphs makes sense.

\begin{figure}
    \centering
    \includegraphics[width=\linewidth]{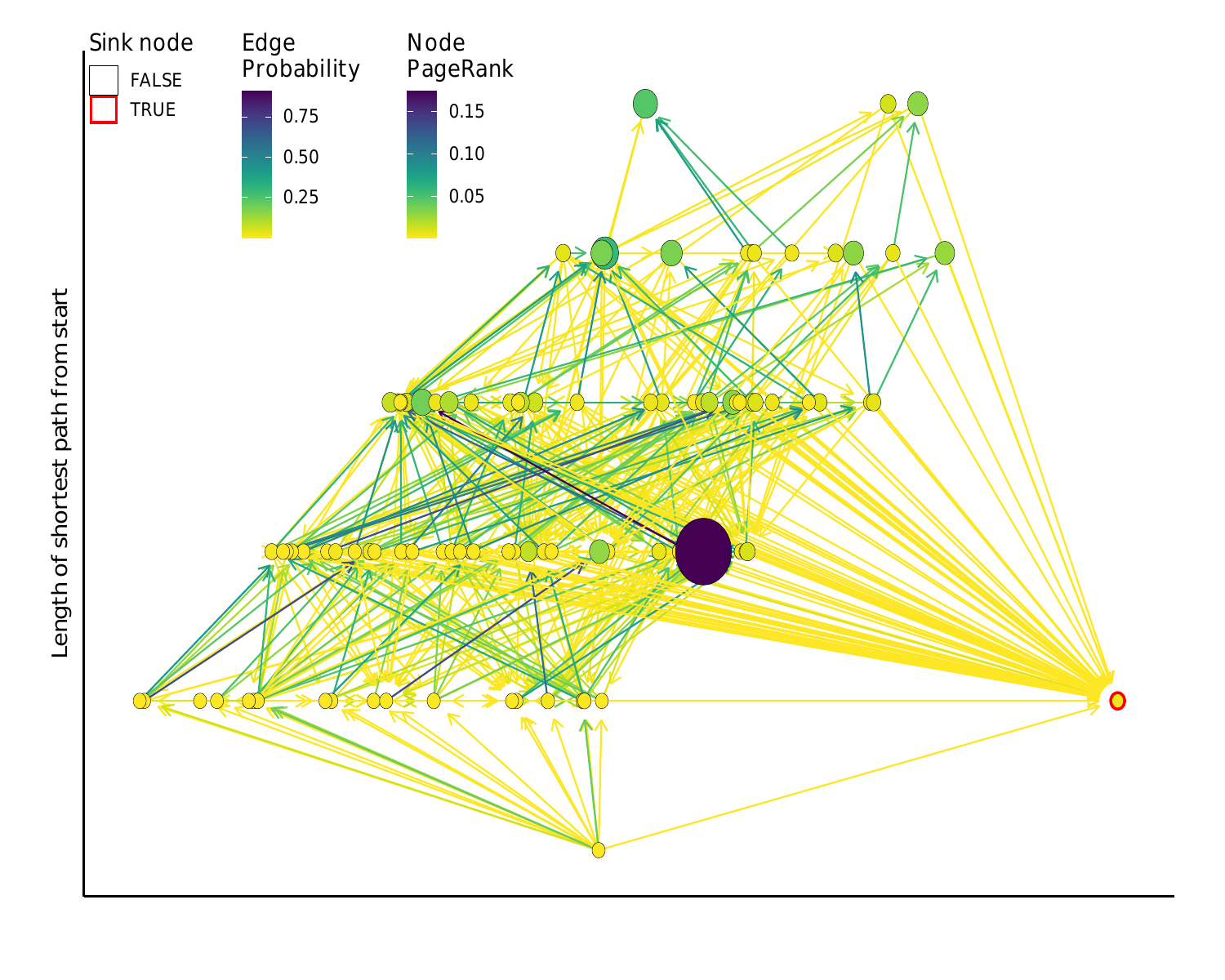}
    \caption{\textbf{Community assembly graph of the 100 most accessible communities for the FizzBuzz problem.} Node colors and sizes indicate the PageRank of each node. Edge colors indicate the probability of choosing each edge. Nodes are arranged along the y axis according to how far away from the starting node they are (measured as shortest path). The starting node (representing a community containing only the worst-performing phenotype) is the lowest node on the y axis. Note that this portion of the graph contains only one true sink node (outlined in red). In this case, that node represents the optimal solution, indicating that the solution for this problem is indeed reachable. However, this node has very low PageRank, indicating that reaching it is relatively unlikely.}
    \label{fig:fizzbuzz}
\end{figure}

Next, we construct a community assembly graph for the FizzBuzz problem~\citep{helmuth_psb2_2021} (see Figure \ref{fig:fizzbuzz}). Interestingly, the optimal solution is also the only true\footnote{Note that there are other nodes that have no outgoing edges within this subgraph; however, these nodes all have outgoing edges to nodes that did not make it into the subgraph because reaching them is unlikely.} sink node in this graph, and it is also technically reachable from the starting community. However, the PageRank for the optimal community is very low, indicating that actually arriving there is unlikely. Instead, there appears to be a node in the middle of the graph that functions as some sort of attractor. Thus, we can conclude that the reason Lalejini et. al found that this problem was harder to solve~\citep{lalejini2023phylogenyinformed} was likely due primarily to mutations producing the optimal genotype being rare and/or evolution spending most of its time in the vicinity of the attractor.

\begin{figure}
    \centering
    \includegraphics[width=\linewidth]{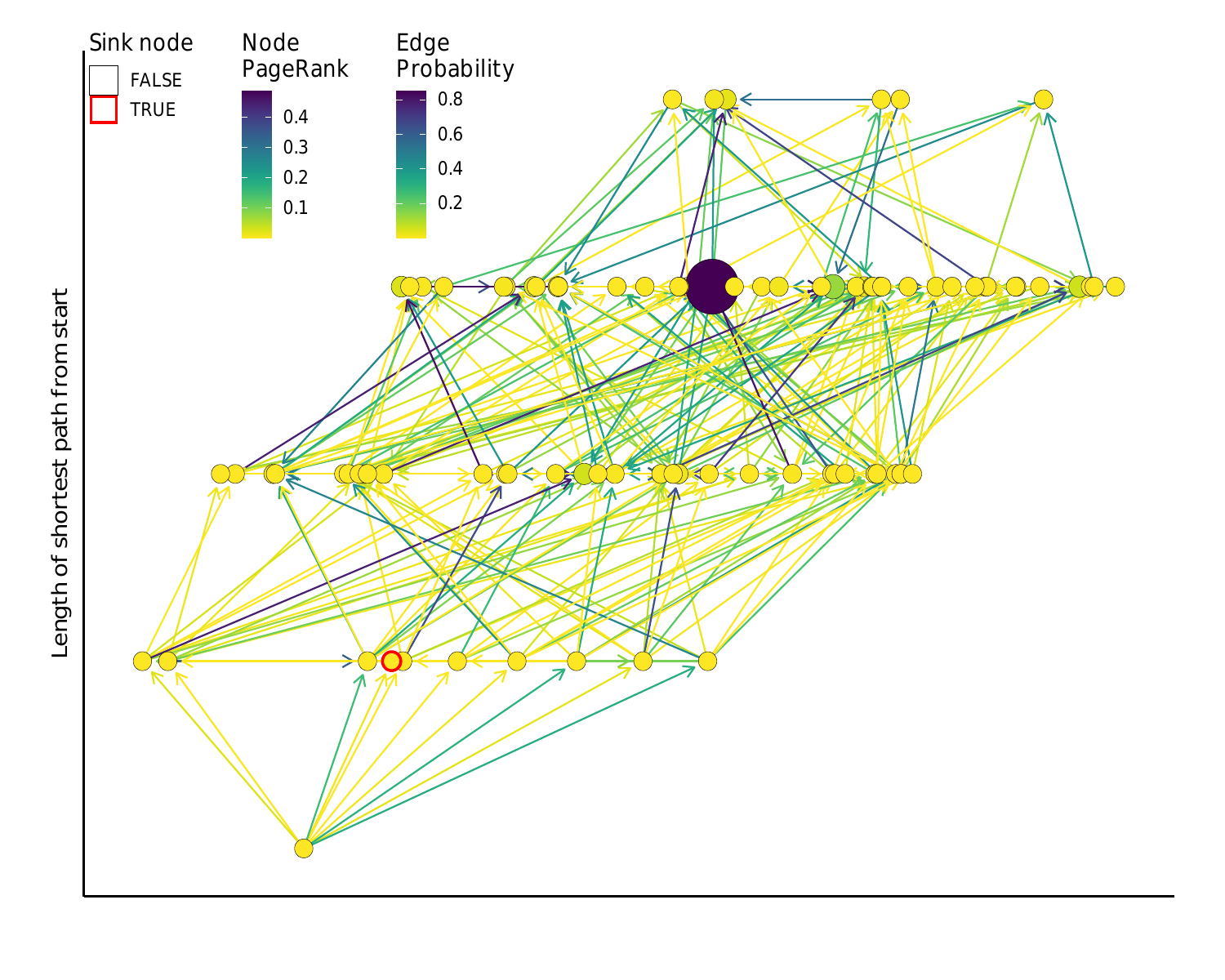}
        \includegraphics[width=\linewidth]{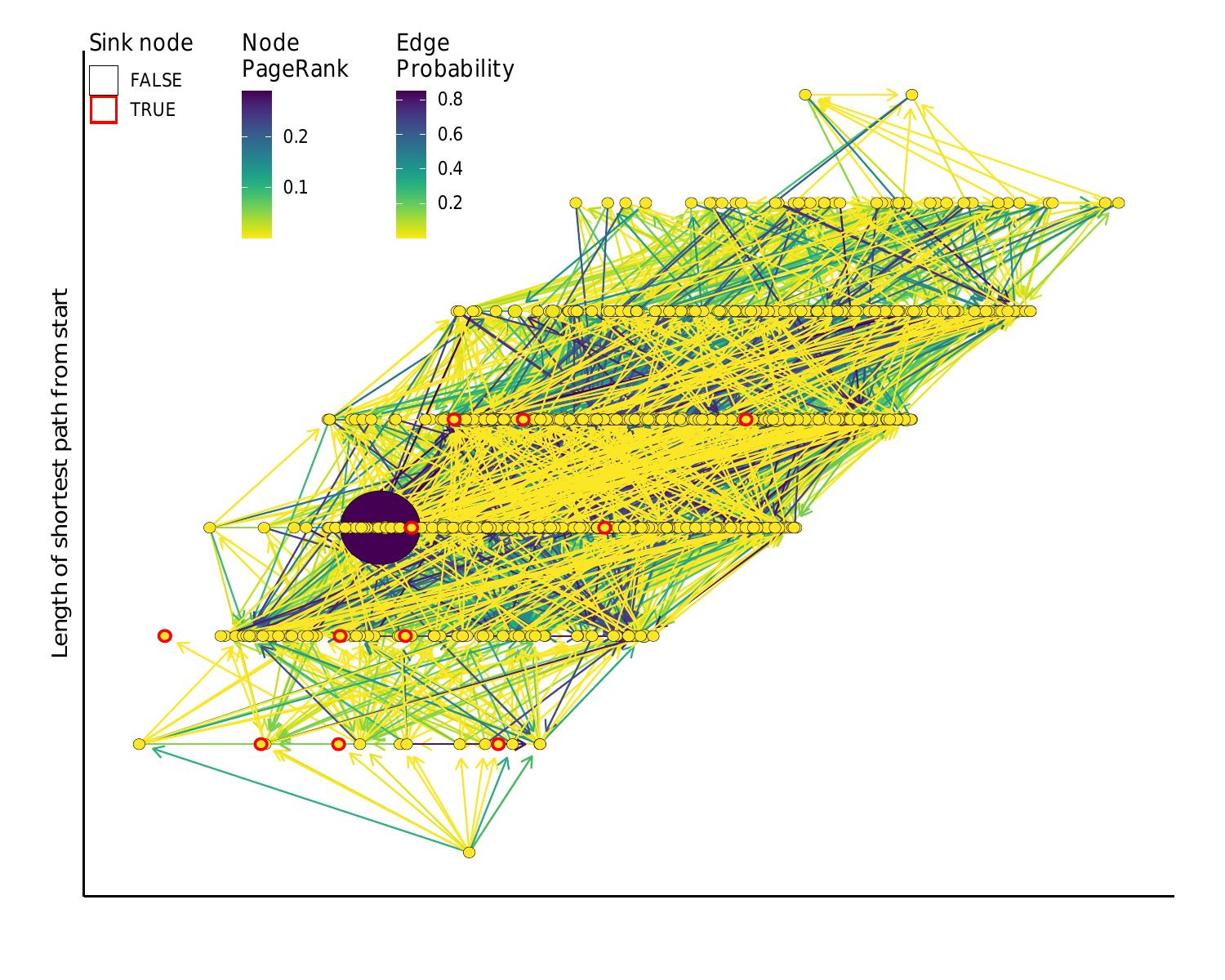}
    \caption{\textbf{Community assembly graph of the 100 (top) and 1000 (bottom) most accessible communities for the small-or-large problem.} Node colors and sizes indicate the PageRank of each node. Edge colors indicate the probability of choosing each edge. The starting node is the lowest node on the y axis. These portions of the graph contain true sink nodes (outlined in red), but they do not correspond to the optimal solution.}
    
    \label{fig:smallorlarge100}
\end{figure}

The small-or-large problem~\citep{helmuth_general_2015} was solved the least frequently in Lalejini et. al's analysis~\citep{lalejini2023phylogenyinformed}. Indeed, none of the 10 runs of evolution used to build our community assembly graph found a solution. Consequently, the optimal solution does not appear in the community assembly graph. There is a true sink node, but it is not easy to reach (see Figure \ref{fig:smallorlarge100} (top)). To understand the impact of subgraph size, we also construct a 1000-node community assembly graph for this problem (see Figure \ref{fig:smallorlarge100} (bottom)). The larger graph has a number of additional sink nodes, although they all still have fairly low PageRanks. Thus, it seems unlikely that there is a single node where the search process is consistently getting stuck. However, some of the difficulty of this problem may arise from there being many different places where it is possible to get stuck.

One important caveat to this analysis is that it is inherently biased by the trajectory that evolution actually took in the underlying runs of evolution. Since we only carried out mutational landscaping around observed genotypes, there are many parts of the fitness landscape that we are unaware of. If we have no examples of runs where a problem was successfully solved, we are very unlikely to conclude that a solution is reachable. However, we believe that this analysis is nevertheless useful for identifying where and why evolution tends to get stuck. 

\section{Conclusion}

Based on our observations thus far, community assembly graphs appear to accurately predict the possible end states of lexicase selection. Consequently, we can use them to identify circumstances where optimal solutions are inaccessible. Moreover, we can use them in the same way that fitness landscapes are used for other evolutionary algorithms: as a tool for understanding unexpected results. Like fitness landscapes, community assembly graphs can be large and costly to calculate. Nevertheless, their ability to precisely calculate possible evolutionary trajectories makes them a powerful analytical tool to have in our tool-kit. In particular, we hypothesize that they may be valuable for comparing the effect of subtle changes to a selection scheme.

Further research is necessary to understand how accurately community assembly graphs built from a small sample of runs of evolution can predict the possible outcomes of subsequent runs. Relatedly, it will be important to understand the impact of sub-graph size on this prediction accuracy. While even the small graphs shown here seem to intuitively describe different evolutionary scenarios, we have also presented preliminary evidence that some dynamics may be missed by small sub-graphs.

A number of additional open questions relate to the interaction between neutral networks and community assembly graphs. Neutral networks are networks of mutationally-adjacent genotypes with the same phenotype. Our analysis assumes that these networks are somewhat sizeable and thus take some time to traverse before a new phenotype can be discovered. This assumption is why we expect it to be uncommon that multiple new phenotypes are introduced to a community at once. Recent analysis by Banzhaf et. al lends preliminary support to this assumption in the context of genetic programming, but raises an important additional layer of nuance: the expected size of neutral networks should decrease with the complexity of program outputs \cite{banzhaf_2023_neutrality}. This pattern likely has implications for the topology of community assembly graphs, although it is not immediately obvious what they are.

% Multiple open questions remain about the extent to which it is safe to make simplifying assumptions for the purposes of constructing community assembly graphs. Preliminary evidence suggests that when these assumptions are violated, lexicase selection may perform more consistently well than the community assembly graph would predict. Determining the importance of rare events such as many-step mutations and the simultaneous introduction of multiple new solutions will be a topic for follow-up research.

In the future, we plan to extend this community assembly graph approach to analyze other evolutionary algorithms in which the probability of various solutions being selected is strongly impacted by the composition of the population. We also plan to develop better techniques for quantitatively analyzing the resulting graphs, as such techniques will enable us to make use of larger sub-graphs than those shown here.

\bibliographystyle{splncs04}
\bibliography{dolson}

\begin{thebibliography}{10}
\providecommand{\url}[1]{\texttt{#1}}
\providecommand{\urlprefix}{URL }
\providecommand{\doi}[1]{https://doi.org/#1}

\bibitem{banzhaf_2023_neutrality}
Banzhaf, W., Hu, T., Ochoa, G.: How the combinatorics of neutral spaces leads
  gp to discover simple solutions (2023), to appear in Genetic Programming
  Theory and Practice XX.

\bibitem{boldiInformedDownSampledLexicase2023}
Boldi, R., Briesch, M., Sobania, D., Lalejini, A., Helmuth, T., Rothlauf, F.,
  Ofria, C., Spector, L.: Informed down-sampled lexicase selection: Identifying
  productive training cases for efficient problem solving (2023).
  \doi{10.48550/arXiv.2301.01488}, \url{http://arxiv.org/abs/2301.01488}

\bibitem{capitanStatisticalMechanicsEcosystem2009}
Capitán, J.A., Cuesta, J.A., Bascompte, J.: Statistical mechanics of ecosystem
  assembly. Physical Review Letters  \textbf{103}(16),  168101 (2009).
  \doi{10.1103/PhysRevLett.103.168101},
  \url{https://link.aps.org/doi/10.1103/PhysRevLett.103.168101}, publisher:
  American Physical Society

\bibitem{ding_optimizing_2022}
Ding, L., Spector, L.: Optimizing {Neural} {Networks} with {Gradient}
  {Lexicase} {Selection}. In: International {Conference} on {Learning}
  {Representations} (2022), \url{https://openreview.net/forum?id=J_2xNmVcY4}

\bibitem{dolsonCalculatingLexicaseSelection2023}
Dolson, E.: Calculating lexicase selection probabilities is np-hard. In:
  Proceedings of the Genetic and Evolutionary Computation Conference. p.
  1575–1583. GECCO '23, Association for Computing Machinery, New York, NY,
  USA (2023). \doi{10.1145/3583131.3590356},
  \url{https://doi.org/10.1145/3583131.3590356}

\bibitem{dolsonApplyingEcologicalPrinciples2018}
Dolson, E., Banzhaf, W., Ofria, C.: Applying {{Ecological Principles}} to
  {{Genetic Programming}}. In: Banzhaf, W., Olson, R.S., Tozier, W., Riolo, R.
  (eds.) Genetic {{Programming Theory}} and {{Practice XV}}, pp. 73--88.
  {Springer International Publishing}, {Cham} (2018)

\bibitem{dolson_phylotrackpy_2023}
Dolson, E., Rodriguez-Papa, S., Moreno, M.A.: Phylotrack: C++ and python
  libraries for in silico phylogenetic tracking. Journal of Open Source
  Software  (in review). \doi{10.5281/zenodo.7922092},
  \url{https://doi.org/10.5281/zenodo.7922092}

\bibitem{dolson_ecological_2018}
Dolson, E.L., Banzhaf, W., Ofria, C.: Ecological theory provides insights about
  evolutionary computation. PeerJ Preprints  \textbf{6},  e27315v1 (Nov 2018).
  \doi{10.7287/peerj.preprints.27315v1}

\bibitem{hang1993assembly}
Hang-Kwang, L., Pimm, S.L.: The assembly of ecological communities: a
  minimalist approach. Journal of Animal Ecology pp. 749--765 (1993)

\bibitem{hang-kwangAssemblyEcologicalCommunities1993}
Hang-Kwang, L., Pimm, S.L.: The assembly of ecological communities: A
  minimalist approach. Journal of Animal Ecology  \textbf{62}(4),  749--765
  (1993). \doi{10.2307/5394}, \url{https://www.jstor.org/stable/5394},
  publisher: [Wiley, British Ecological Society]

\bibitem{helmuth_psb2_2021}
Helmuth, T., Kelly, P.: {PSB2}: the second program synthesis benchmark suite.
  In: Proceedings of the {Genetic} and {Evolutionary} {Computation}
  {Conference}. pp. 785--794. ACM, Lille France (Jun 2021).
  \doi{10.1145/3449639.3459285},
  \url{https://dl.acm.org/doi/10.1145/3449639.3459285}

\bibitem{helmuth_general_2015}
Helmuth, T., Spector, L.: General {Program} {Synthesis} {Benchmark} {Suite}.
  In: Proceedings of the 2015 on {Genetic} and {Evolutionary} {Computation}
  {Conference} - {GECCO} '15. pp. 1039--1046. ACM Press, Madrid, Spain (2015).
  \doi{10.1145/2739480.2754769},
  \url{http://dl.acm.org/citation.cfm?doid=2739480.2754769}

\bibitem{hernandezRandomSubsamplingImproves2019}
Hernandez, J.G., Lalejini, A., Dolson, E., Ofria, C.: Random subsampling
  improves performance in lexicase selection. In: Proceedings of the Genetic
  and Evolutionary Computation Conference Companion. pp. 2028--2031. {GECCO}
  '19, Association for Computing Machinery (2019).
  \doi{10.1145/3319619.3326900}, \url{https://doi.org/10.1145/3319619.3326900}

\bibitem{hernandez_suite_2022}
Hernandez, J.G., Lalejini, A., Ofria, C.: A suite of diagnostic metrics for
  characterizing selection schemes (Sep 2022). \doi{10.48550/arXiv.2204.13839}

\bibitem{kauffmanGeneralTheoryAdaptive1987}
Kauffman, S., Levin, S.: Towards a general theory of adaptive walks on rugged
  landscapes. Journal of Theoretical Biology  \textbf{128}(1),  11--45 (1987).
  \doi{10.1016/S0022-5193(87)80029-2},
  \url{http://www.sciencedirect.com/science/article/pii/S0022519387800292}

\bibitem{lacavaProbabilisticMultiObjectiveAnalysis2018}
La~Cava, W., Helmuth, T., Spector, L., Moore, J.H.: A {{Probabilistic}} and
  {{Multi-Objective Analysis}} of {{Lexicase Selection}} and
  {$\epsilon$}-{{Lexicase Selection}}. Evolutionary Computation pp. 1--26 (May
  2018). \doi{10.1162/evco\_a\_00224}

\bibitem{la_cava_epsilon-lexicase_2016}
La~Cava, W., Spector, L., Danai, K.: Epsilon-{Lexicase} {Selection} for
  {Regression}. In: Proceedings of the {Genetic} and {Evolutionary}
  {Computation} {Conference} 2016. pp. 741--748. ACM, Denver Colorado USA (Jul
  2016). \doi{10.1145/2908812.2908898},
  \url{https://dl.acm.org/doi/10.1145/2908812.2908898}

\bibitem{lacavaEpsilonLexicaseSelectionRegression2016}
La~Cava, W., Spector, L., Danai, K.: Epsilon-{{Lexicase Selection}} for
  {{Regression}}. In: Proceedings of the {{Genetic}} and {{Evolutionary
  Computation Conference}} 2016. pp. 741--748. {{GECCO}} '16, {Association for
  Computing Machinery}, {New York, NY, USA} (Jul 2016).
  \doi{10.1145/2908812.2908898}

\bibitem{lalejini_artificial_2022}
Lalejini, A., Dolson, E., Vostinar, A.E., Zaman, L.: Artificial selection
  methods from evolutionary computing show promise for directed evolution of
  microbes. eLife  \textbf{11},  e79665 (Aug 2022). \doi{10.7554/eLife.79665},
  \url{https://elifesciences.org/articles/79665}

\bibitem{lalejini2023phylogenyinformed}
Lalejini, A., Moreno, M.A., Hernandez, J.G., Dolson, E.: Phylogeny-informed
  fitness estimation (2023), to appear in Genetic Programming Theory and
  Practice XX.

\bibitem{lalejini_evolving_2018}
Lalejini, A., Ofria, C.: Evolving event-driven programs with {SignalGP}. In:
  Proceedings of the {Genetic} and {Evolutionary} {Computation} {Conference} on
  - {GECCO} '18. pp. 1135--1142. ACM Press, Kyoto, Japan (2018).
  \doi{10.1145/3205455.3205523},
  \url{http://dl.acm.org/citation.cfm?doid=3205455.3205523}

\bibitem{pappa_faster_2023}
Matsumoto, N., Saini, A.K., Ribeiro, P., Choi, H., Orlenko, A., Lyytikäinen,
  L.P., Laurikka, J.O., Lehtimäki, T., Batista, S., Moore, J.H.: Faster
  {Convergence} with {Lexicase} {Selection} in {Tree}-{Based} {Automated}
  {Machine} {Learning}. In: Pappa, G., Giacobini, M., Vasicek, Z. (eds.)
  Genetic {Programming}, vol. 13986, pp. 165--181. Springer Nature Switzerland,
  Cham (2023). \doi{10.1007/978-3-031-29573-7_11},
  \url{https://link.springer.com/10.1007/978-3-031-29573-7_11}, series Title:
  Lecture Notes in Computer Science

\bibitem{metevier_lexicase_2019}
Metevier, B., Saini, A.K., Spector, L.: Lexicase {Selection} {Beyond} {Genetic}
  {Programming}. In: Banzhaf, W., Spector, L., Sheneman, L. (eds.) Genetic
  {Programming} {Theory} and {Practice} {XVI}, pp. 123--136. Springer
  International Publishing, Cham (2019). \doi{10.1007/978-3-030-04735-1_7},
  \url{http://link.springer.com/10.1007/978-3-030-04735-1_7}, series Title:
  Genetic and Evolutionary Computation

\bibitem{moore_lexicase_2017}
Moore, J.M., Stanton, A.: Lexicase selection outperforms previous strategies
  for incremental evolution of virtual creature controllers. In: Proceedings of
  the 14th {European} {Conference} on {Artificial} {Life} {ECAL} 2017. pp.
  290--297. MIT Press, Lyon, France (Sep 2017). \doi{10.7551/ecal_a_050},
  \url{https://www.mitpressjournals.org/doi/abs/10.1162/isal_a_050}

\bibitem{schreiberSimpleRulesCycling2004}
Schreiber, S.J., Rittenhouse, S.: From simple rules to cycling in community
  assembly. Oikos  \textbf{105}(2),  349--358 (2004).
  \doi{10.1111/j.0030-1299.2004.12433.x},
  \url{https://onlinelibrary.wiley.com/doi/abs/10.1111/j.0030-1299.2004.12433.x},
  \_eprint:
  https://onlinelibrary.wiley.com/doi/pdf/10.1111/j.0030-1299.2004.12433.x

\bibitem{servan2021tractable}
Serv{\'a}n, C.A., Allesina, S.: Tractable models of ecological assembly.
  Ecology Letters  \textbf{24}(5),  1029--1037 (2021)

\bibitem{spectorAssessmentProblemModality2012}
Spector, L.: Assessment of problem modality by differential performance of
  lexicase selection in genetic programming: a preliminary report. In:
  Proceedings of the 14th annual conference companion on Genetic and
  evolutionary computation. pp. 401--408. {ACM} (2012),
  \url{http://dl.acm.org/citation.cfm?id=2330846}

\bibitem{spectorRelaxationsLexicaseParent2018}
Spector, L., Cava, W.L., Shanabrook, S., Helmuth, T., Pantridge, E.:
  Relaxations of lexicase parent selection. In: Banzhaf, W., Olson, R.S.,
  Tozier, W., Riolo, R. (eds.) Genetic Programming Theory and Practice {XV}.
  pp. 105--120. Genetic and Evolutionary Computation, Springer International
  Publishing (2018). \doi{10.1007/978-3-319-90512-9\_7}

\bibitem{ostmanPredictingEvolutionVisualizing2014}
Østman, B., Adami, C.: Predicting evolution and visualizing high-dimensional
  fitness landscapes. In: Recent Advances in the Theory and Application of
  Fitness Landscapes, pp. 509--526. Emergence, Complexity and Computation,
  Springer, Berlin, Heidelberg (2014). \doi{10.1007/978-3-642-41888-4_18},
  \url{https://link.springer.com/chapter/10.1007/978-3-642-41888-4_18}

\end{thebibliography}

\end{document}